\documentclass[11pt,a4paper]{article}
\usepackage[hyperref]{ranlp2025}

\usepackage{times}
\usepackage{latexsym}

\usepackage{lipsum}
\usepackage{array}
\usepackage{amsmath}
\usepackage{amsfonts}
\usepackage{graphicx}
\usepackage{microtype}
\newcommand{\footURL}[1]{\footnote{\url{#1}}}
\usepackage{subcaption}
\aclfinalcopy 

\usepackage{ifthen}
\newcommand{\footL}[2][]{\footnote{\scriptsize \urlstyle{tt}\url{#2}\ifthenelse {\equal {#1} {}}{}{\label{note:#1}}}}

\title{
Zero-shot OCR Accuracy of Low-Resourced Languages: A Comparative Analysis on Sinhala and Tamil
}

\author{Nevidu Jayatilleke \and Nisansa de Silva \\
  Department of Computer Science \& Engineering, \\
  University of Moratuwa, Sri Lanka \\
  \texttt{\{nevidu.25, NisansaDdS\}@cse.mrt.ac.lk} 
  }

\date{}

\begin{document}
\maketitle
\begin{abstract}
Solving the problem of Optical Character Recognition (OCR) on printed text for Latin and its derivative scripts can now be considered settled due to the volumes of research done on English and other High-Resourced Languages (HRL). However, for Low-Resourced Languages (LRL) that use unique scripts, it remains an open problem. This study presents a comparative analysis of the zero-shot performance of six distinct OCR engines on two LRLs: Sinhala and Tamil.
%
The selected engines include both commercial and open-source systems, aiming to evaluate the strengths of each category. 
The Cloud Vision API, Surya, Document AI, and Tesseract were evaluated for both Sinhala and Tamil, while Subasa OCR and EasyOCR were examined for only one language due to their limitations. The performance of these systems was rigorously analysed using five measurement techniques to assess accuracy at both the character and word levels. According to the findings,
Surya delivered the best performance for Sinhala across all metrics, with a WER of 2.61\%.
Conversely, Document AI excelled across all metrics for Tamil, highlighted by a very low CER of 0.78\%.
%
In addition to the above analysis, we also introduce a novel synthetic Tamil OCR benchmarking dataset\footL[tamilocrdata]{https://huggingface.co/datasets/Nevidu/tamil_synthetic_ocr}.
\end{abstract}
\section{Introduction}

Optical Character Recognition (OCR) is a computational technology that is used for recognising text within digital images, such as scanned documents, advertisements, and photographs~\cite{agarwal-anastasopoulos-2024-concise, jain2021ocr, weerasinghe-etal-2008-nlp}. 
OCR is commonly employed as an information entry tool to extract valuable data from scanned documents such as forms, receipts, invoices, and passports. Historically, OCR (along with Text-To-Speech systems) was created to assist blind or disabled individuals by facilitating machines to read written text aloud to them, a development that dates back to 1914~\cite{mittal2020text}.

The process of OCR typically involves multiple steps: 1) It begins with image acquisition, where the image is captured. 2) Next is pre-processing, which enhances the image quality and includes binarisation to separate the content from the background. 3) Following this is layout analysis, where the document is divided into distinct regions. 4) The next step is character-level segmentation, which breaks the text down into lines, words, and individual characters. 5) Recognition follows, involving feature extraction and classification to identify the characters. 6) Finally, post-processing improves the results, often using language models. Each of these stages is essential for effective OCR performance~\cite{jain2021ocr, nazeem-etal-2024-open}.

While OCR systems have advanced significantly, particularly for High-Resourced Languages (HRL) such as English and French~\cite{nazeem-etal-2024-open}, recognising text from complex or low-quality images, historical documents, and Low-Resource Languages (LRL) still presents challenges~\cite{agarwal-anastasopoulos-2024-concise}. In this benchmarking study, we conduct a thorough examination of various multilingual and monolingual OCR systems, assessing their capabilities for two selected low-resource languages in South Asia: Sinhala and Tamil. 

\begin{figure}[!htbp]
    \centering
    \setlength{\fboxsep}{1pt} 
    \setlength{\fboxrule}{0.4pt} 
    \fbox{%
        \includegraphics[width=0.8\columnwidth]{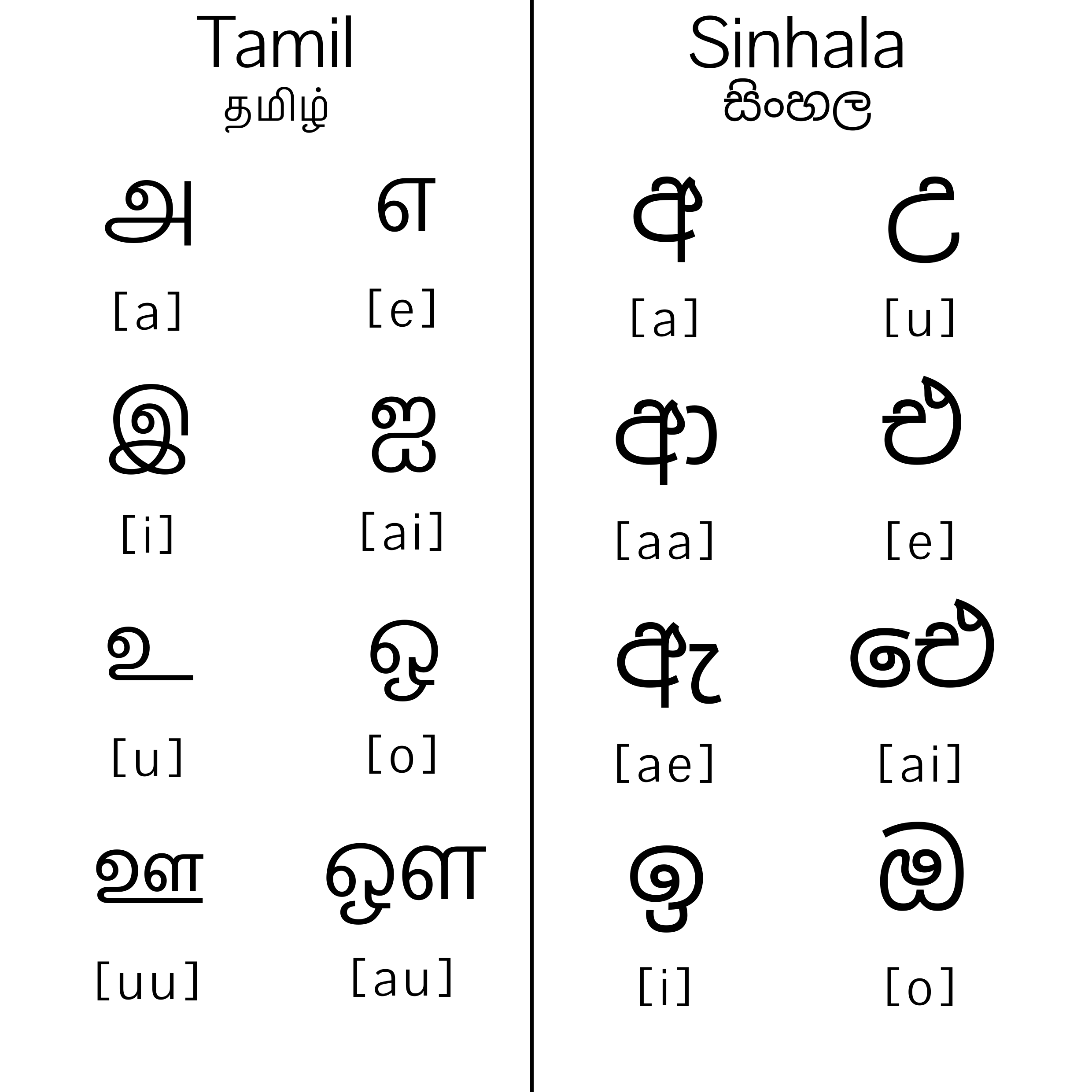}%
    }%
    \caption{An example of the use of rounded script in Tamil and Sinhala languages.}
    \label{fig:roundedscripts}
\end{figure}

Sinhala is an Indo-European language spoken as L1 by just 16 million people, mostly located in the island of Sri Lanka~\cite{de2025survey}. Sinhala has a script that is unique to it which descends from the Indian Brahmi Script~\cite{fernando1949palaeographical}. 
Tamil is a Dravidian language spoken as L1 by around 79 million people located primarily in India, Sri Lanka, and Singapore~\cite{wijeratne2019natural}. It also has a unique script that is also a descendant of the Indian Brahmi Script~\cite{paneerselvam1972critical}.
Both Sinhala and Tamil are considered LRLs by the criteria proposed by~\citet{ranathunga-de-silva-2022-languages}, where Sinhala is deemed to be lower resourced (Category 02) as opposed to Tamil (Category 03).

\section{Existing Works}


Despite extensive research over the past several decades, the challenge of accurately recognising Sinhala characters in OCR systems remains a formidable obstacle~\cite{anuradha2020deep}. Indic languages, such as Tamil, present a myriad of complexities and character variations, significantly complicating the development of effective OCR solutions. 
In clear terms, the accuracy of South Asian rounded scripts is significantly behind that of Latin-based scripts, highlighting a crucial area for improvement and development~\cite{anuradha2021estimating}. 

\subsection{Sinhala OCR Systems}

Several research studies have been conducted on developing Sinhala OCR systems. 
A system utilising the Tesseract 4.0 OCR engine with a graphical user interface has been proposed for the Sinhala language by~\citet{anuradha2020deep}. The system comprises five main components: the user, an Application Programming Interface (API), the Tesseract engine~\cite{smith2007overview}, a post-processor, and a data store. The Tesseract engine processes images and recognises text using Long Short Term Memory (LSTM) based deep learning techniques. However, some characters are not recognised by the Tesseract engine. To address this issue, the post-processor identifies those unrecognised characters and applies linguistic rules to ensure accurate output. For this study, commercially used font types were employed, varying in size, which resulted in an average accuracy of 94\%. 

A study on multi-style printed Sinhala character recognition~\cite{maduranga2022multi} utilised a hybrid Artificial Neural Network (ANN) model, incorporating concepts from previous research. The process includes four steps: Data Preprocessing for image enhancement and noise removal; Feature Extraction, dividing 50x50px character images into 9 zones with 12 pieces each to create a feature vector of 108 signals for the ANN; Development and Training, using line features from an 850-character database (primarily Iskoola Pota font\footL[iskoolapota]{https://learn.microsoft.com/en-us/typography/font-list/iskoola-pota}) to train a backpropagation network in MATLAB with 108 input nodes, 78 hidden nodes, and 34 output nodes over 138 epochs, achieving around 75\% training accuracy; and Testing, where a separate dataset of 1253 characters is used to evaluate performance.



\citet{velayuthan2025benchmarking} conducted a comparative analysis addressing the challenge of digitising documents in low-resource languages by benchmarking several prominent OCR models: Surya\footL[suryaocr]{https://github.com/VikParuchuri/surya}, TR-OCR\footL[tr-ocr]{https://huggingface.co/Ransaka/TrOCR-Sinhala}, EasyOCR\footL[easyocr]{https://github.com/JaidedAI/EasyOCR}, and Tesseract OCR~\cite{smith2007overview}. While it initially states a focus on Sinhala and Tamil, the research ultimately conducted experiments and reported results only for Sinhala using two synthetic Sinhala datasets, as well as for English using the FUNSD dataset~\cite{jaume2019funsd}. The evaluation process, employing metrics such as CER (Character Error Rate) and WER (Word Error Rate), involved data filtration and post-processing techniques. The findings revealed that Surya significantly outperformed the other models on the Sinhala datasets. 
Although all models demonstrated high error rates on the English dataset, Surya was identified as the most balanced option for Sinhala, achieving relatively high accuracy while keeping moderate computational demands. Additionally, Surya demonstrated superior power efficiency, using an average of 0.69 kWh less across the two Sinhala datasets compared to TR-OCR. The accuracy difference between Sinhala and English is largely due to the datasets used; Sinhala datasets are synthetically generated, while the English dataset consists of noisy form document images.


\subsection{Tamil OCR Systems}

Various research initiatives have been undertaken to create OCR systems for the Tamil language. Unlike Sinhala, which is primarily spoken in Sri Lanka, the Tamil language has a much wider geographical spread across the South Asian region. 
Notwithstanding their different linguistic roots, Tamil also utilises a rounded script similar to Sinhala, as depicted in Figure~\ref{fig:roundedscripts}, due to their writing systems being related. This leads to similar challenges in the development of OCR technology. 

\citet{liyanage2015developing} developed a Tamil OCR system using the open-source Tesseract OCR engine, inspired by its applications with scripts such as Sinhala and Bangla. The methodology involved creating a 169-character OCR alphabet and preparing training data from selected words in various Unicode fonts. They tested different training combinations and found that a model using data from three fonts at three sizes achieved the best results. The system was evaluated using 20 scanned images from ancient Tamil books, achieving an accuracy of 81\%. This was benchmarked only against Tesseract's existing Tamil module, over which, a 12.5\% improvement was shown.

Recent research introduced the \textit{Nayana framework}~\cite{kolavi-etal-2025-nayana}, which enhances Vision-Language Models (VLMs) such as GOT (General OCR Theory)~\cite{wei2024general} for low-resource languages, including Tamil. It tackles data scarcity through a layout-aware synthetic data generation pipeline and Low-Rank Adaptation (LoRA)~\cite{hu2022lora}. This system translates English documents into Tamil while retaining their layout, followed by a two-phase Cross-Modal Alignment training with LoRA. Nayana-OCR achieved notable improvements in performance, with a Word Error Rate (WER) of 0.551 and a Metric for Evaluation of Translation with Explicit ORdering (METEOR) score of 0.592, significantly outperforming the base GOT model (WER 1.020, METEOR 0.051) and other traditional OCR systems, including Tesseract~\cite{smith2007overview} and PaddleOCR\footL{https://github.com/PaddlePaddle/PaddleOCR}, on the Tamil test set.


\section{Methodology}
\label{sec:length}

As discussed earlier, contemporary studies are utilising open-source tools to effectively fine-tune models for new languages and improve existing capabilities. Many of these tools offer multilingual support. Additionally, several organisations have developed commercial engines that excel in OCR. In this study, we thoroughly evaluate the capabilities of selected OCR engines for Sinhala and Tamil languages in a zero-shot setting.

\subsection{Overview of Selected OCR Technologies}

In this study, we benchmark the capabilities of six selected open-source and commercial engines specifically designed for OCR tasks.

\noindent\textbf{Cloud Vision API\footL{https://cloud.google.com/vision/docs/ocr}:} The Cloud Vision API enables developers to seamlessly incorporate vision detection features into their applications. This includes functionalities such as image labelling, face and landmark detection, OCR, and the tagging of explicit content. The first version of the API launched for general availability in May 2017. The API is designed to perform OCR on files such as Portable Document Formats (PDFs) and Tag Image File Formats (TIFFs), as well as on images with dense text. It is particularly optimised for image documents containing large amounts of text and those that feature handwriting, allowing for accurate recognition and conversion to machine-readable text.

\noindent\textbf{Document AI\footL{https://cloud.google.com/document-ai/}:} Document AI is a document understanding platform that transforms unstructured data from documents into structured data, making it easier to understand, analyse, and utilise. It employs machine learning and Google Cloud to develop scalable, end-to-end cloud-based document processing applications. The API provides organisation through content classification, entity extraction, advanced searching, and more. The OCR processor specifically enables the identification and extraction of text, including handwritten text, from documents in over 200 languages. Additionally, the processor uses machine learning to assess the quality of a document based on the readability of its content.

\noindent\textbf{Tesseract:} One of the most well-known and frequently used OCR engines is Tesseract OCR. This open-source project, originally developed by Hewlett-Packard (HP) and now funded by Google, offers excellent text recognition capabilities. Tesseract combines Hidden Markov Models (HMMs) and various machine learning algorithms with traditional computer vision techniques to recognise text. The Tesseract 4.0 incorporates deep learning methods that significantly improve performance compared to earlier versions that primarily relied on conventional approaches. The deep learning models utilised by Tesseract are based on LSTM networks \cite{smith2007overview, nazeem-etal-2024-open}. In this study, we utilise Tesseract 5.5.0, an improved version of Tesseract 4.0, which incorporates the existing LSTM engine, along with several performance improvements.

\noindent\textbf{Subasa OCR\footL[subsaocr]{https://ocr.subasa.lk/}:} The study by~\citet{anuradha2020deep} was extended, which involved a comprehensive series of deep learning (LSTM) based Tesseract 4.0 experiments to estimate the complexity of Sinhala OCR by investigating the effects of text genre, image resolution, and algorithmic complexity. Training data was prepared primarily from the UCSC 10M Sinhala corpus\footL[ucsc10m]{https://ltrl.ucsc.lk/tools-and-resourses/}, using various Sinhala fonts and image qualities, involving character segmentation and iterative training processes. Evaluation was conducted on 30 non-identical test images categorised as old newspapers (200 DPI), old books (72 DPI), and contemporary books (300 DPI), with an additional test on low-DPI (96px) contemporary images. Accuracy was the primary metric, calculated by comparing character counts between the original text and the OCR output. This OCR engine achieved up to 67.02\% character accuracy on old newspapers, up to 87.53\% on old books, and up to 87.63\% on contemporary books, even maintaining high accuracy up to 87.88\% on low-DPI contemporary images \cite{anuradha2021estimating}.

\noindent\textbf{Surya\footref{note:suryaocr}:} This is an OCR toolkit that supports over 90 languages and benchmarks favourably against cloud services. It features line-level text detection in any language, layout analysis (including detection of tables, images, headers, etc.), reading order detection, and table recognition (detecting rows and columns), as well as LaTeX OCR capabilities. The text detection model was trained on four A6000 GPUs for three days using a diverse set of images. It was built from the ground up using a modified EfficientViT architecture~\cite{liu2023efficientvit} for semantic segmentation. Meanwhile, the text recognition model was trained on the same hardware for two weeks, utilising a modified Donut model~\cite{kim2022ocr} that incorporates Grouped Query Attention (GQA)~\cite{ainslie-etal-2023-gqa}, a Mixture of Experts (MoE) layer~\cite{shazeeroutrageously}, Unicode Transformation Format-16 (UTF-16) decoding, and changes to the layer configuration. It is important to note that this system is designed for printed text and not for handwriting.

\noindent\textbf{EasyOCR\footref{note:easyocr}:} This is an OCR technique that supports over 80 languages. EasyOCR utilises ResNet \cite{he2016deep}, LSTM, and Connectionist Temporal Classification (CTC)~\cite{graves2006connectionist} models for character recognition. The detection component of EasyOCR employs the Character Region Awareness For Text detection (CRAFT) Algorithm~\cite{baek2019character}. EasyOCR consists of three key elements. The first is feature extraction, which is executed by the ResNet model. The second element is sequence labelling, for which the LSTM algorithm is utilised, and the last component is decoding which relies on CTC. The EasyOCR's \texttt{Readtext} function is utilized during text recognition which can read letters and numbers from images while providing their location coordinates~\cite{awalgaonkar2021automatic}.

The Google Cloud Vision API and Document AI are commercial engines, whereas Google Tesseract, Surya, and EasyOCR are open-source systems. Additionally, we selected Subasa OCR, which is a fine-tuned version of the Tesseract model available through a web application\footref{note:subsaocr}, though the source code and model are not directly accessible.

\subsection{Dataset Selection and Assembly}

As noted by~\citet{ranathunga-etal-2024-shoulders,de2025survey}, free and open datasets for this task are scarce for low-resourced languages; even when some published research may exist.
To achieve optimal results, we employed distinct datasets for each of the two selected languages, ensuring a tailored approach that enhances the effectiveness of our analysis. For the Sinhala language, we chose a dataset published in Hugging Face, \texttt{sinhala\_synthetic\_ocr-large}\footL[sinhala-synthetic-ocr-large]{https://huggingface.co/datasets/Ransaka/sinhala_synthetic_ocr-large} by~\citet{ransaka_ravihara_2024}, consisting of 6,969 pairs of images and reference texts created using five different font families; 

\begin{itemize}
\setlength\itemsep{-0.3em}
\item Noto Sans Sinhala\footL[font-noto-sans-sinhala]{https://fonts.google.com/noto/specimen/Noto+Sans+Sinhala}
\item Gemunu Libre\footL[font-gemunu-libre]{https://fonts.google.com/specimen/Gemunu+Libre}
\item Noto Serif Sinhala\footL[font-noto-serif-sinhala]{https://fonts.google.com/noto/specimen/Noto+Serif+Sinhala}
\item Yaldevi\footL[font-yaldevi]{https://fonts.google.com/specimen/Yaldevi}
\item Abhaya Libre\footL[font-abhaya-libre]{https://fonts.google.com/specimen/Abhaya+Libre}
\end{itemize}

\begin{figure}[htbp!] 
    \centering
    \includegraphics[width=0.7\columnwidth]{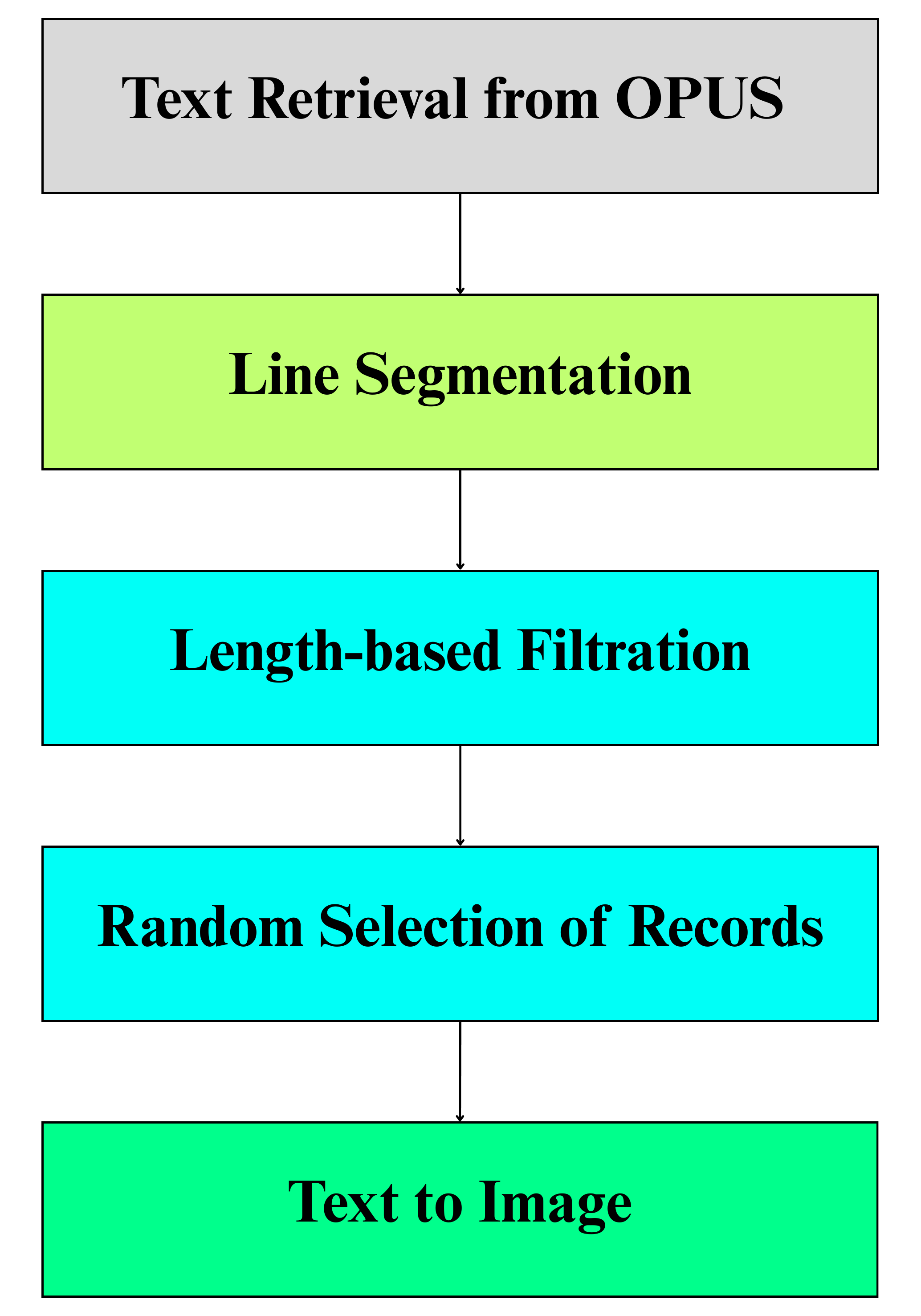} 
    \caption{Overview of the Tamil synthetic OCR dataset creation.}
    \label{fig:tamil_synthetic_ocr}
\end{figure}

Since we considered a synthetically generated dataset for Sinhala, we also aimed to evaluate synthetically generated data for Tamil to ensure a fair comparison. However, we could not find any publicly available Tamil datasets developed in a similar manner to the selected Sinhala dataset. As a result, we decided to create a new dataset for the Tamil language. The overview of the Tamil dataset creation is shown in Figure~\ref{fig:tamil_synthetic_ocr}.

The Tamil text was obtained from OPUS\footL[opus]{https://opus.nlpl.eu/}~\cite{tiedemann-2012-parallel}, specifically selecting OpenSubtitles~\cite{lison-tiedemann-2016-opensubtitles2016} v2024\footL[opus-opensub]{https://opus.nlpl.eu/OpenSubtitles/ta&en/v2024/OpenSubtitles}. The content was then divided line by line, focusing solely on Tamil characters since the primary goal was language evaluation, resulting in 2,437,960 records. Subsequently, this set was filtered to retain only texts longer than 40 characters, yielding 222,658 records. This filtration step ensured that word-level evaluation was accurate. However, to ensure a fair comparison with the Sinhala dataset, we decided to randomly select 7,000 samples from the remaining texts to equalise the sample sizes. This step was necessary to prevent evaluation scores from being skewed by differences in dataset size. Further, conducting OCR on additional samples is challenging due to the resource consumption involved.

We carefully selected six unique font families out of a total of 17 from Google Fonts\footL[google-fonts]{https://fonts.google.com/} to diversify the creation of images from the text records, ensuring that each visual representation is impactful for the analysis. We identified the unique fonts visually, as some fonts, such as Mukta Malar\footL[mukta-malar]{https://fonts.google.com/specimen/Mukta+Malar} and Baloo Thambi 2\footL[baloo-thambi]{https://fonts.google.com/specimen/Baloo+Thambi+2}, have characters that can appear very similar to the naked eye. The selected fonts are as follows:

\begin{itemize}
\setlength\itemsep{-0.3em}
\item Hind Madurai\footL[font-hind-madurai]{https://fonts.google.com/specimen/Hind+Madurai}
\item Noto Serif Tamil\footL[font-noto-serif-tamil]{https://fonts.google.com/noto/specimen/Noto+Serif+Tamil}
\item Kavivanar\footL[font-kavivanar]{https://fonts.google.com/specimen/Kavivanar}
\item Noto Sans Tamil\footL[font-noto-sans-tamil]{https://fonts.google.com/noto/specimen/Noto+Sans+Tamil}
\item Pavanam\footL[font-pavanam]{https://fonts.google.com/specimen/Pavanam}
\item Anek Tamil\footL[font-anek-tamil]{https://fonts.google.com/specimen/Anek+Tamil}
\end{itemize}

A function was then developed utilising the capabilities of the \texttt{Pillow}\footL[pillow]{https://pillow.readthedocs.io/} library to systematically convert textual data into image files. Its fundamental purpose is to ensure a proportional distribution of input text records across the defined set of font files, promoting an equitable use of fonts in the generated image dataset. For each text entry, the function dynamically calculates optimal image dimensions based on the measurements of the text bounding box, rendering the text in black on a white background. A significant feature of this implementation is the precise centring of the text within the generated image, achieved through calculated positioning in relation to the height and width of the image, thereby enhancing visual consistency and quality.

\begin{figure*}[htbp!] 
    \centering

    \setlength{\fboxsep}{0.5pt} 
    \setlength{\fboxrule}{0.8pt}
    \fbox{%
    \includegraphics[width=1.0\textwidth]{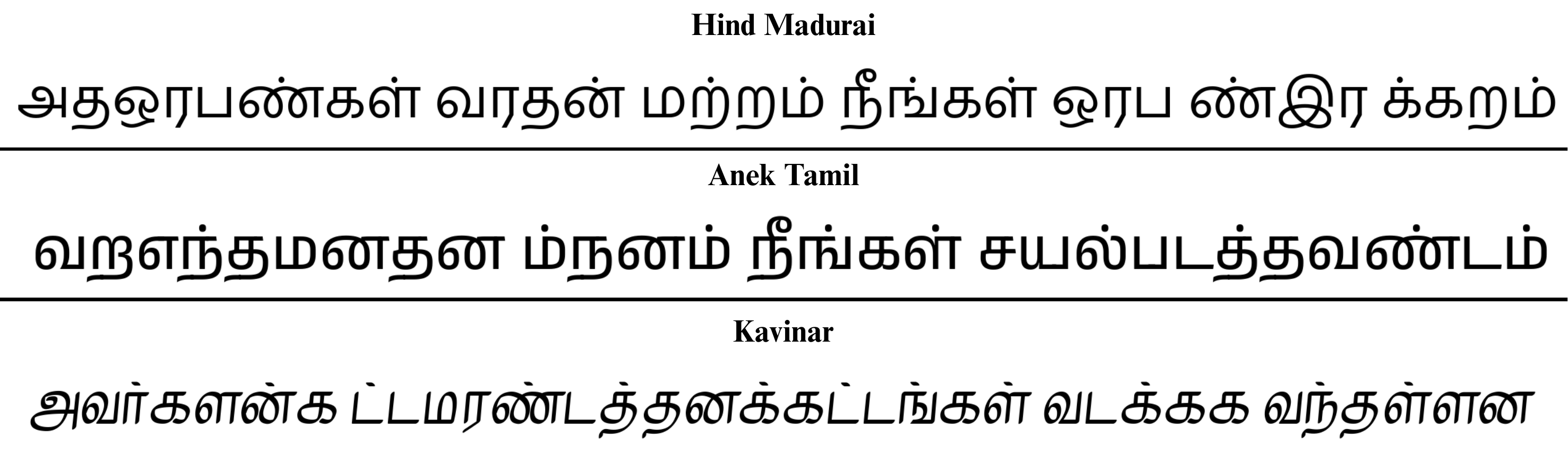} 
    }%
    \caption{Three examples of Tamil sentences from our dataset in the fonts Hind Madurai, Anek Tamil, and Kavinar, respectively.}
    \label{fig:tamil_sentences}
\end{figure*}

The post-processing phase for the Tamil dataset involved the exclusion of records that contained no string values for both reference and generated features. In contrast, the post-processing for the Sinhala dataset incorporated an additional step aimed at eliminating all non-Sinhala characters from both the reference and generated texts. This step was implemented to ensure that the evaluation concentrated exclusively on the targeted linguistic capabilities of the system. Since we created the Tamil dataset, the postprocessing step focused on the language characters was addressed during preprocessing. This novel synthetic Tamil OCR public dataset is one of our contributions in this study\footref{note:tamilocrdata}.
Some examples of Tamil sentences from our synthetic dataset are shown in Figure~\ref{fig:tamil_sentences}.

\subsection{Integration of OCR Systems}

Both the Google Cloud Vision API and Document AI are available through the Google Cloud Platform (GCP). However, while enabling the Cloud Vision API is straightforward, as it only requires activating the API in GCP, Document AI necessitates the manual creation of a processor within GCP. This created processor is then used to initiate the OCR process.

The process of integrating the Tesseract engine was quite simple through the use of the \texttt{pytesseract}\footL[pytesseract]{https://pypi.org/project/pytesseract/} wrapper. This enabled the automation of character recognition for every record within each dataset, streamlining the overall data processing workflow.

As previously mentioned, the Subasa OCR engine is exclusively accessible via a web application. This limitation meant that we had to manually input images one by one to perform OCR, a process that proves impractical given our substantial Sinhala dataset of 6,969 records. To streamline this tedious task and enhance efficiency, we resolved to automate it using \texttt{Selenium}\footL[selenium]{https://www.selenium.dev/} after meticulously identifying the element locations by inspecting the web source code. 

The seamless integration of Surya and EasyOCR was straightforward, largely due to their comprehensive documentation. Both engines are accessible on GitHub and can be conveniently installed as libraries directly through Python, making the setup process efficient and user-friendly.

\subsection{Evaluation Mechanism}

The evaluation was performed using the generated text against the reference text of the datasets. For the comparison, we employed five different measurements;


\noindent\textbf{Character Error Rate (CER):} It is based on the concept of Levenshtein distance, which measures the minimum number of character-level operations (substitutions, deletions, and insertions) necessary to transform the ground truth text into the output generated by OCR. The CER formula is expressed as $(S + I + D) / N$, where $S$ represents the number of substitutions, $I$ denotes insertions, $D$ signifies deletions, and $N$ is the total number of characters in the ground truth text~\cite{nazeem-etal-2024-open}. 

\noindent\textbf{Word Error Rate (WER):} Similar to the CER, the WER is calculated by comparing the text produced by the OCR system against a ground truth or reference text. The WER is determined by the number of word-level errors made by the OCR engine. The formula for calculating the word error rate is also $(S + I + D) / N$, but considered at the word level instead of the character level 
\cite{nazeem-etal-2024-open}.

\noindent\textbf{Bilingual Evaluation Understudy (BLEU):} It is a method created for the automatic evaluation of machine translation quality. The fundamental principle is that the closer a machine-generated translation is to one or more professional human translations, the higher its quality. BLEU evaluates this proximity through a numerical metric that relies on a corpus of high-quality human reference translations. The methodology involves a weighted average of variable-length phrase matches against these reference translations, utilising a concept known as modified n-gram precision. Additionally, it incorporates a brevity penalty to discourage candidate translations that are overly brief in comparison to the reference translations. The final BLEU score, which ranges from 0 to 1, is derived by calculating the geometric mean of the modified n-gram precisions and multiplying this by the brevity penalty 
\cite{papineni-etal-2002-bleu}.

\noindent\textbf{Average Normalised Levenshtein Similarity (ANLS):} This metric takes into account both reasoning errors and shortcomings of OCR. To evaluate answers, it calculates a similarity score between the response of the model and the ground truth using Levenshtein distance. A key feature of this scoring system is its application of a threshold ($\tau$=0.5) on the normalised Levenshtein distance (NL): if NL is less than or equal to 0.5, the similarity score is calculated as 1-NL; otherwise, the score is 0. This approach allows ANLS to provide intermediate scores (ranging from 0.5 to 1) for responses that are logically correct but may contain minor recognition errors, contrasting with standard accuracy metrics, which would score them as zero \cite{biten2019scene}.

\begin{equation}
\text{ANLS} = \frac{1}{N} \sum_{i=0}^{N} \left( \max_{j} s(a_{ij}, o_{q_i}) \right)
\end{equation}

\begin{equation*} 
\small 
s(a_{ij}, o_{q_i}) =
\begin{cases}
1 - NL(a_{ij}, o_{q_i}) & \text{if } NL(a_{ij}, o_{q_i}) < \tau \\
0 & \text{if } NL(a_{ij}, o_{q_i}) \geq \tau
\end{cases}
\end{equation*}

\noindent\textbf{Metric for Evaluation of Translation with Explicit ORdering (METEOR):} Similar to BLEU, METEOR is also a metric designed for evaluating the quality of machine translation. It operates by identifying generalised unigram matches between machine-generated outputs and human reference translations, allowing for matches based on surface forms, stemmed forms, and synonyms. Designed specifically to address limitations of the BLEU metric, such as its inability to directly account for recall and its indirect measurement of word order, METEOR computes a score by integrating unigram precision, unigram recall (with greater emphasis on recall), and a fragmentation penalty that evaluates the word order of matched words. This method has demonstrated improved correlation with human judgments when compared to other evaluation techniques. Given its enhanced ability to assess generated text against references and its rapid adoption in OCR studies, we chose to utilise this metric~\cite{banerjee-lavie-2005-meteor}.


\section{Discussion of Results}

\begin{table*}[!htb]
\centering
\begin{tabular}{llccccc}
\hline
\textbf{OCR System} & \textbf{Language} & \textbf{CER}$\downarrow$ & \textbf{WER}$\downarrow$ & \textbf{BLEU}$\uparrow$ & \textbf{ANLS}$\uparrow$ & \textbf{METEOR}$\uparrow$ \\
\hline
Cloud Vision API & Sinhala & 0.0619 & 0.0767 & 0.9193 & 0.9447 & 0.9269 \\
& Tamil & 0.0079 & 0.1204 & 0.5790 & 0.9922 & 0.8751 \\
\hline
Surya & Sinhala & \textbf{0.0076} & \textbf{0.0261} & \textbf{0.9396} & \textbf{0.9920} & \textbf{0.9723} \\
& Tamil & 0.1392 & 0.6500 & 0.1487 & 0.8672 & 0.3359 \\
\hline
Document AI & Sinhala & 0.0610 & 0.0758 & 0.9199 & 0.9455 & 0.9278 \\
& Tamil & \textbf{0.0078} & \textbf{0.1198} & \textbf{0.5803} & \textbf{0.9923} & \textbf{0.8762} \\
\hline
Subasa OCR & Sinhala & 0.0761 & 0.1799 & 0.6894 & 0.9259 & 0.8099 \\
& Tamil & - & - & - & - & - \\
\hline
Tesseract & Sinhala & 0.0702 & 0.1489 & 0.7553 & 0.9319 & 0.8436 \\
& Tamil & 0.0780 & 0.6145 & 0.0493 & 0.9264 & 0.3201 \\
\hline
EasyOCR & Sinhala & - & - & - & - & - \\
& Tamil & 0.1172 & 0.2876 & 0.3461 & 0.8828 & 0.6744 \\
\hline
\end{tabular}
\caption{\label{tab:results}
The evaluation of OCR systems for the Sinhala and Tamil languages 
}
\end{table*}

\newcommand{\radialFigSize}{0.4\columnwidth}

\begin{figure*}[htbp!]
    \centering
    \begin{subfigure}[t]{\radialFigSize} 
        \centering
        \includegraphics[width=\columnwidth]{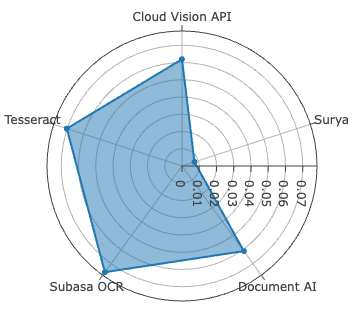} 
        \caption{CER-Sinhala$\downarrow$}
    \end{subfigure}
    \begin{subfigure}[t]{\radialFigSize} 
        \centering
\includegraphics[width=\columnwidth]{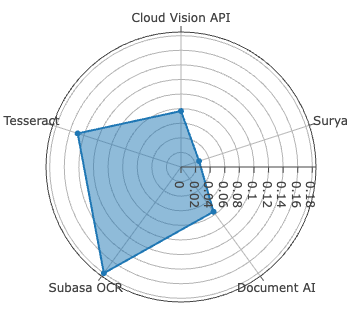} 
        \caption{WER-Sinhala$\downarrow$}
    \end{subfigure}
        \begin{subfigure}[t]{\radialFigSize} 
        \centering
        \includegraphics[width=\columnwidth]{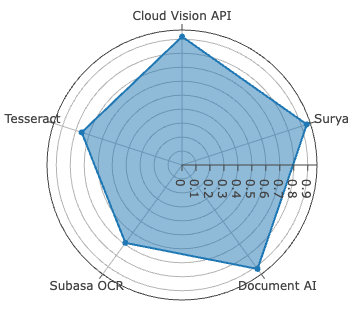} 
        \caption{BLEU-Sinhala$\uparrow$}
    \end{subfigure}
        \begin{subfigure}[t]{\radialFigSize} 
        \centering
        \includegraphics[width=\columnwidth]{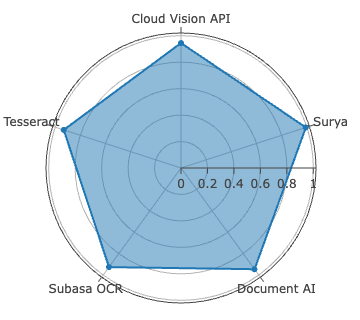} 
        \caption{ANLS-Sinhala$\uparrow$}
    \end{subfigure}
    \begin{subfigure}[t]{\radialFigSize} 
        \centering
        \includegraphics[width=\columnwidth]{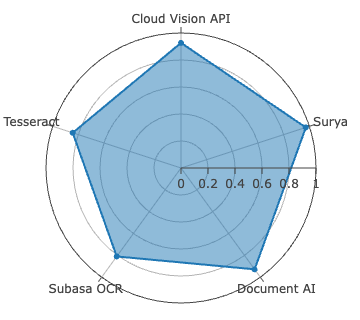} 
        \caption{METEOR-Sinhala$\uparrow$}
    \end{subfigure}

    \begin{subfigure}[t]{\radialFigSize} 
        \centering
        \includegraphics[width=\columnwidth]{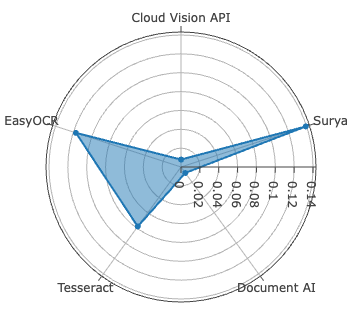} 
        \caption{CER-Tamil$\downarrow$}
    \end{subfigure}  
    \begin{subfigure}[t]{\radialFigSize} 
        \centering
        \includegraphics[width=\columnwidth]{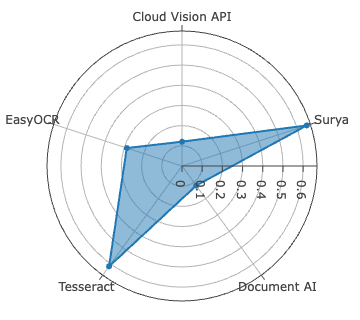} 
        \caption{WER-Tamil$\downarrow$}
    \end{subfigure}
    \begin{subfigure}[t]{\radialFigSize} 
        \centering
        \includegraphics[width=\columnwidth]{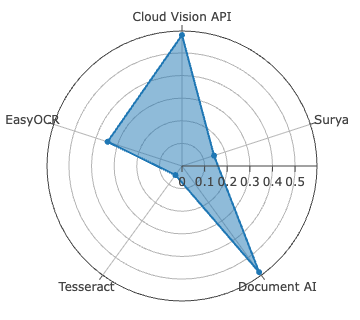} 
        \caption{BLEU-Tamil$\uparrow$}
    \end{subfigure}
    \begin{subfigure}[t]{\radialFigSize} 
        \centering
        \includegraphics[width=\columnwidth]{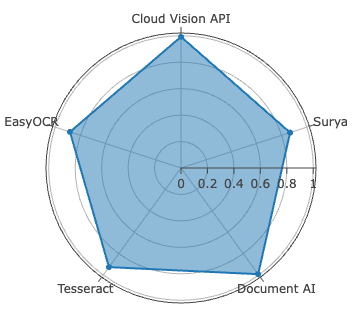} 
        \caption{ANLS-Tamil$\uparrow$}
    \end{subfigure}
    \begin{subfigure}[t]{\radialFigSize} 
        \centering
        \includegraphics[width=\columnwidth]{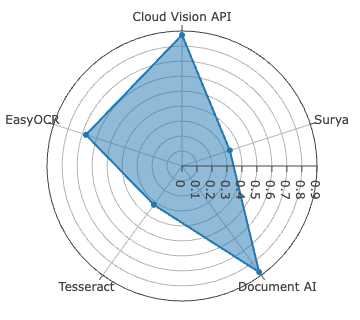} 
        \caption{METEOR-Tamil$\uparrow$}
    \end{subfigure}

    \caption{Results for Sinhala and Tamil}
    \label{fig:WERresults}
\end{figure*}

In this section, we present the analysis of the results drawn from our comparative evaluation of OCR systems. The findings are showcased distinctly for both Sinhala and Tamil in Table~\ref{tab:results}.
With the exception of Subasa OCR and EasyOCR, all four other systems were able to 
process both languages selected for evaluation in this study. It is important to note that Subasa OCR is a monolingual system specifically fine-tuned for Sinhala. While EasyOCR has multilingual capabilities, it notably lacks support for the Sinhala language.
When evaluating the overall results for both languages, the Cloud Vision API and Document AI have 
achieved very similar results. However, Document AI outperforms the other engines for Tamil across all metrics. 

Overall, the best WER results for Tamil are notably higher than those for Sinhala.
This observation highlights the disparity between character and word identification. Document AI, despite achieving a very low CER, achieves a comparatively higher WER, indicating that while the system effectively identifies characters, it struggles with word formation and spacing of the Tamil language. This issue is common across all engines and applies to both languages, as systems tend to perform better at recognising characters but struggle with forming words and managing spacing. In contrast, the METEOR and ANLS scores for both languages are relatively high, suggesting a strong alignment in terms of content, word order, and semantic meaning. However, the BLEU scores for Tamil are markedly lower than those of other metrics, likely due to the elevated WER, which results in fewer successful n-gram overlaps.

Performance of Surya on the Sinhala language has been nothing short of extraordinary, emerging as the standout among the others. The metrics clearly illustrate this success, as highlighted in Table~\ref{tab:results}. When we compare the best WER for Tamil at 11.98\% with that of Sinhala at an impressive 2.61\% as depicted in Figure~\ref{fig:WERresults}, the superiority of the accuracy of the Surya engine for Sinhala becomes strikingly apparent. Furthermore, the METEOR and ANLS scores of 0.9723 and 0.9920, respectively, underscore its near-perfect word-level performance. 

    

The comparison between Subasa OCR and Tesseract is particularly compelling, as Subasa OCR represents a fine-tuned adaptation of the Tesseract 4.0 engine specifically for Sinhala. The authors of Subasa assert that their modifications yield significantly superior results compared to the standard Tesseract 4.0~\cite{anuradha2021estimating}. The metric evaluations reveal that Tesseract 5.5.0 now outperforms Subasa OCR across all metrics. This indicates that the latest version of Tesseract by Google has made substantial enhancements for the Sinhala language, even in its vanilla form. However, Tesseract's performance in Tamil is not competitive compared to other systems.

As discussed earlier, due to its lack of support for Sinhala, EasyOCR is evaluated
solely on Tamil, on which it demonstrated superior performance among the open-source systems we compared. While the results show a notable decline compared to two commercial engines, the contrast with other open-source solutions is significant. 
%
%

Furthermore, we performed a character-level error analysis for the best models of each language. This analysis involved counting the number of errors per character by comparing the generated text to a reference. In Sinhala, we identified erroneous characters based on a threshold of more than 5,000 errors. Diacritics such as `\raisebox{-0.3ex}{ 
    \includegraphics[height=1.0\fontcharht\font`\A]{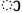} 
}', `\raisebox{-0.9ex}{ 
    \includegraphics[height=1.5\fontcharht\font`\A]{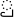} 
}', `\raisebox{-0.3ex}{ 
    \includegraphics[height=1.5\fontcharht\font`\A]{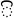} 
}', and `\raisebox{-0.3ex}{ 
    \includegraphics[height=1.6\fontcharht\font`\A]{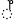} 
}' were particularly difficult to identify. In addition, letters such as `\raisebox{-0.1ex}{ 
    \includegraphics[height=0.8\fontcharht\font`\A]{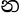} 
}', `\raisebox{-0.1ex}{ 
    \includegraphics[height=0.8\fontcharht\font`\A]{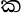} 
}', `\raisebox{-0.1ex}{ 
    \includegraphics[height=0.8\fontcharht\font`\A]{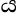} 
}', `\raisebox{-0.1ex}{ 
    \includegraphics[height=1.0\fontcharht\font`\A]{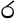} 
}', and `\raisebox{-0.1ex}{ 
    \includegraphics[height=1.0\fontcharht\font`\A]{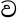} 
}' posed challenges for Surya in terms of accurate detection. 

For Tamil, we set a threshold of more than 1,600 errors to identify erroneous characters. The diacritic `\raisebox{-0.3ex}{ 
    \includegraphics[height=1.5\fontcharht\font`\A]{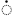} 
}' emerged as the most error-prone character, while letters such as `\raisebox{-0.1ex}{ 
    \includegraphics[height=0.8\fontcharht\font`\A]{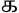} 
}', `\raisebox{-0.5ex}{ 
    \includegraphics[height=1.1\fontcharht\font`\A]{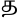} 
}', `\raisebox{-0.1ex}{ 
    \includegraphics[height=0.8\fontcharht\font`\A]{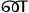} 
}', and `\raisebox{-0.5ex}{ 
    \includegraphics[height=1.1\fontcharht\font`\A]{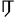} 
}' were also among the most problematic for Document AI.

The number of character errors is generally higher for Sinhala compared to Tamil. However, the best CER scores for both languages do not highlight this difference. This discrepancy arises because, despite having a greater number of character-level errors, the overall usage frequency per character in Sinhala is also comparably higher, which reduces the average error rate. This phenomenon is potentially due to the differences in the sizes of their alphabets\footnote{The moniker \textit{Alphabet} is used in the general meaning here. Both these scripts are in fact \textit{Abugidas} rather than \textit{Alphabets}.}: Sinhala has 60 characters, while Tamil contains 247.

\section{Conclusion}

In this study, we evaluated six different OCR engines for two distinct South Asian languages in a zero-shot setting. To facilitate this evaluation, we created a synthetic Tamil OCR dataset, utilising six different fonts to be parallel to the existing Sinhala dataset. 
The performance of selected OCR systems was thoroughly analysed using five measurements that evaluated accuracy levels at both the character and word levels. 

The results indicated that Document AI performed best for Tamil, while Surya excelled in Sinhala. Both the Cloud Vision API and Document AI showed reasonable performance in OCR for these languages, highlighting the capabilities of commercial engines, as anticipated.
A standout performer was Surya for Sinhala, which outperformed all other OCR systems in each metric. Furthermore, the significant disparity between the best CER and WER results for Tamil compared to Sinhala indicates that while Document AI excels at character recognition, it falls short in accurately identifying words through proper character formation and white-space detection. Additionally, zero-shot Tesseract 5.5.0 outperformed a fine-tuned Tesseract 4.0 system on Sinhala (Subasa OCR). Moreover, The differences in scores between the commercial OCR systems are largely a black box, likely arising from nuances in their architectures and training data.

Out of approximately 2.2 million Tamil text records, we selected only 7,000 records to ensure a fair comparison with the Sinhala dataset, which has only 6,969 records. In future studies, it would be possible to expand the two synthetic datasets and consider more fonts, backgrounds, and varied noise conditions to create more realistic simulations.


\section*{Limitations}

The analysis compared the performance of various OCR engines on Tamil and Sinhala printed scripts. Although fine-tuning the systems has the potential to improve performance, it was not carried out as this study intended only to analyse existing systems. It is also important to note a limitation regarding the datasets used; both are synthetically created, featuring black text on a white background. This design results in clean and clear images, which do not accurately reflect real-world conditions when capturing printed text. The decision to use synthetic data was made because similar realistic datasets were not available for the two low-resourced languages, and it was necessary to ensure a fair comparison.

While a considerable amount of work exists for Indic languages such as Hindi, according to the observations of~\citet{de2025survey}, the Devanagari script of Hindi has a distance of 5 from Sinhala and a distance of 7 from Tamil, as opposed to a distance of 4 between Sinhala and Tamil. Further, according to~\citet{how2021Rao}, scripts such as Sinhala and Tamil are considered rounded scripts, while Hindi is not. For these reasons, our work could not rely much on the progress made for Hindi OCR. 

With real-world data, OCR technology faces significant challenges due to input image quality, particularly with historical documents and low-resource language data. Factors such as poor print quality, low resolution, shading, blurriness, and distortion can severely impact accuracy~\cite{hegghammer2022ocr}. Images that exhibit distortion, textured backgrounds, cluttered environments, disconnected line segments, isolated dots, breaks in lines, rotation, motion blur, and out-of-focus blur complicate character segmentation and recognition, often leading to higher error rates~\cite{karunarathne2017recognizing}. Additionally, low image resolutions can slow down OCR speed due to greater uncertainty in character representation~\cite{anuradha2021estimating}.
Consequently, when evaluated using camera-captured images, the accuracy levels may notably differ from the results presented in this study.

\bibliographystyle{acl_natbib}
\bibliography{anthology,ranlp2025}

\end{document}